# Exact Soft Confidence-Weighted Learning

Jialei Wang                                                        JL.WANG@NTU.EDU.SG
Peilin Zhao                                                        ZHAO0106@NTU.EDU.SG
Steven C.H. Hoi                                                    CHHOI@NTU.EDU.SG
School of Computer Engineering, Nanyang Technological University, Singapore 639798

## Abstract

In this paper, we propose a new Soft Confidence-Weighted (SCW) online learning scheme, which enables the conventional confidence-weighted learning method to handle non-separable cases. Unlike the previous confidence-weighted learning algorithms, the proposed soft confidence-weighted learning method enjoys all the four salient properties: (i) large margin training, (ii) confidence weighting, (iii) capability to handle non-separable data, and (iv) adaptive margin. Our experimental results show that the proposed SCW algorithms significantly outperform the original CW algorithm. When comparing with a variety of state-of-the-art algorithms (including AROW, NAROW and NHERD), we found that SCW generally achieves better or at least comparable predictive accuracy, but enjoys significant advantage of computational efficiency (i.e., smaller number of updates and lower time cost).

## 1. Introduction

Online learning algorithms (Rosenblatt, 1958; Crammer et al., 2006; Jin et al., 2010; Zhao et al., 2011a;b) represent a family of fast and simple machine learning techniques, which usually make few statistical assumptions and can be applied to a wide range of applications (Li et al., 2012). Online learning has been actively studied in machine learning community, in which a variety of online learning algorithms have been proposed, including a number of first-order algorithms such as the well-known Perceptron algorithm (Rosenblatt, 1958) and the Passive-Aggressive (PA) algorithms (Crammer et al., 2006).

Recent years have seen a surge of studies on the second-order online learning algorithms (Cesa-Bianchi et al., 2005; Dredze et al., 2008; Crammer et al., 2009b; Orabona & Crammer, 2010; Duchi et al., 2011), which have shown that parameter confidence information can be explored to guide and improve online learning performance (Cesa-Bianchi et al., 2005). For example, Confidence-weighted (CW) learning (Dredze et al., 2008; Crammer et al., 2009a) maintains a Gaussian distribution over some linear classifier hypotheses and applies it to control the direction and scale of parameter updates (Dredze et al., 2008). Although CW learning has formal guarantees in the mistake-bound model (Crammer et al., 2008), it can overfit in certain situations due to its aggressive update rules based upon a separable data assumption. Recently, an improved online algorithm, i.e., Adaptive Regularization of Weights (AROW) (Crammer et al., 2009b; Orabona & Crammer, 2010), relaxes such separable assumption by employing an adaptive regularization for each training example based upon its current confidence. This regularization comes in the form of minimizing a combination of the Kullback-Leibler divergence between Gaussian distributed weight vectors and a confidence penalty of vectors.

Although AROW is able to improve the original CW learning by handling noisy and non-separable cases, it is not the exact corresponding soft extending part of CW (Like PA with PA-I and PA-II). In particular, the directly added loss and confidence regularization make AROW lose an important property of Confidence-weighted learning, i.e., Adaptive Margin property. Following the similar idea of soft margin support vector machines, the adaptive margin assigns different margins for different instances via a probability formulation, which enables CW to gain extra efficiency and effectiveness.

In this work, we extend the confidence-weighted learning for soft margin learning, which makes our Soft

Confidence-Weighted (SCW) learning method more robust than the original CW learning when handling noisy and non-separable data, and more effective and efficient than the state-of-the-art AROW algorithm.

The rest of this paper is organized as follows. Section 2 reviews related work. Section 3 proposes the soft confidence-weighted learning method. Section 4 analyzes the mistake bounds and properties of our algorithms. Section 5 conducts an extensive set of empirical experiments, and Section 6 concludes this work.

## 2. Related Work and Background

### 2.1. Overview of Online Learning

Online learning operates on a sequence of data examples with time stamps. At time step $t$, the algorithm processes an incoming example $\mathbf{x}_t \in \mathbb{R}^d$ by first predicting its label $\hat{y}_t \in \{-1, +1\}$. After the prediction, the true label $y_t \in \{-1, +1\}$ is revealed and then the loss $\ell(y_t, \hat{y}_t)$, which is the difference between its prediction and the revealed true label $y_t$, is suffered. Finally, the loss is used to update the weights of the model based on some criterion. Overall, the goal of online learning is to minimize the cumulative mistake over the entire sequence of data examples.

Our work is closely related to several first and second order online learning algorithms, including Passive-Aggressive (PA) learning (Crammer et al., 2006), Confidence-Weighted learning (Dredze et al., 2008), and Adaptive Regularization of Weights learning (Crammer et al., 2009b). Below we review the basics of these algorithms.

### 2.2. Passive-Aggressive Learning

As the state-of-the-art first order online learning algorithm, the optimization of Passive-Aggressive (PA) learning is formulated as:

$$\mathbf{w}_{t+1} = \arg\min_{\mathbf{w} \in \mathbb{R}^d} \frac{1}{2}\|\mathbf{w} - \mathbf{w}_t\|^2 \, s.t. \, \ell(\mathbf{w}; (\mathbf{x}_t, y_t)) = 0 \quad (1)$$

where the loss function is based on the hinge loss:

$$\ell(\mathbf{w}; (\mathbf{x}_t, y_t)) = \begin{cases} 0 & \text{if } y_t(\mathbf{w} \cdot \mathbf{x}_t) \geq 1 \\ 1 - y_t(\mathbf{w} \cdot \mathbf{x}_t) & \text{otherwise} \end{cases}$$

The above optimization has the closed-form solution:

$$\mathbf{w}_{t+1} = \mathbf{w}_t + \eta_t^{\text{PA}} y_t \mathbf{x}_t \quad (2)$$

where $\eta_t^{\text{PA}} = \frac{\ell(\mathbf{w}_t;(\mathbf{x}_t,y_t))}{\|\mathbf{x}_t\|^2}$. Further, to let PA be able to handle non-separable instances and more robust, a slack variable $\xi$ was introduced into the optimization (1) using one of two types of penalty: linear and quadratic, leading to the following two formulations of soft-margin PA algorithms:

$$\mathbf{w}_{t+1}^{\text{PA-I}} = \arg\min_{\mathbf{w} \in \mathbb{R}^d} \frac{1}{2}\|\mathbf{w} - \mathbf{w}_t\|^2 + C\ell(\mathbf{w}; (\mathbf{x}_t, y_t))$$

$$\mathbf{w}_{t+1}^{\text{PA-II}} = \arg\min_{\mathbf{w} \in \mathbb{R}^d} \frac{1}{2}\|\mathbf{w} - \mathbf{w}_t\|^2 + C\ell(\mathbf{w}; (\mathbf{x}_t, y_t))^2$$

where $C$ is a parameter to tradeoff between passiveness and aggressiveness. The resulting weight updates to the soft-margin PA algorithms have the same form as that of (2), but different coefficients $\eta_t$ as follows:

$$\eta_t^{\text{PA-I}} = \min\{C, \frac{\ell(\mathbf{w}_t;(\mathbf{x}_t,y_t))}{\|\mathbf{x}_t\|^2}\}, \eta_t^{\text{PA-II}} = \frac{\ell(\mathbf{w}_t;(\mathbf{x}_t,y_t))}{\|\mathbf{x}_t\|^2 + \frac{1}{2C}}$$

### 2.3. Confidence-Weighted Learning

To better exploring the underlying structure between features, the Confidence-Weighted (CW) learning algorithm assumes a Gaussian distribution of weights with mean vector $\boldsymbol{\mu} \in \mathbb{R}^d$ and covariance matrix $\Sigma \in \mathbb{R}^{d \times d}$. The weight distribution is updated by minimizing the Kullback-Leibler divergence between the new weight distribution and the old one while ensuring that the probability of correct classfication is greater than a threshold as follows:

$$(\boldsymbol{\mu}_{t+1}, \Sigma_{t+1}) = \arg\min_{\boldsymbol{\mu},\Sigma} D_{KL}(\mathcal{N}(\boldsymbol{\mu},\Sigma), \mathcal{N}(\boldsymbol{\mu}_t, \Sigma_t))$$

$$s.t. \quad Pr_{\mathbf{w} \sim \mathcal{N}(\boldsymbol{\mu},\Sigma)}[y_t(\mathbf{w} \cdot \mathbf{x}_t) \geq 0] \geq \eta$$

This optimization problem has a closed-form solution

$$\boldsymbol{\mu}_{t+1} = \boldsymbol{\mu}_t + \alpha_t y_t \Sigma_t \mathbf{x}_t \quad \Sigma_{t+1} = \Sigma_t - \beta_t \Sigma_t \mathbf{x}_t^T \mathbf{x}_t \Sigma_t \quad (3)$$

The updating coefficients are calculated as follows:

$$\alpha_t = \max\left\{0, \frac{1}{v_t \zeta}(-m_t \psi + \sqrt{m_t^2 \frac{\phi^4}{4} + v_t \phi^2 \zeta})\right\}$$

$$\beta_t = \frac{\alpha_t \phi}{\sqrt{u_t} + v_t \alpha_t \phi}$$

where $u_t = \frac{1}{4}(-\alpha_t v_t \phi + \sqrt{\alpha_t^2 v_t^2 \phi^2 + 4v_t})^2$, $v_t = \mathbf{x}_t^T \Sigma_t \mathbf{x}_t$, $m_t = y_t(\boldsymbol{\mu}_t \cdot \mathbf{x}_t)$, $\phi = \Phi^{-1}(\eta)$ ($\Phi$ is the cumulative function of the normal distribution), $\psi = 1 + \frac{\phi^2}{2}$, and $\zeta = 1 + \phi^2$.

### 2.4. Adaptive Regularization of Weights

Unlike the original CW learning algorithm, the Adaptive Regularization Of Weights (AROW) learning introduces adaptive regularization of the prediction function when processing a new instance in each learning step, making it more robust than CW to sudden changes of label noise in the learning tasks. In particular, the optimization of AROW is formulated as:

$$(\boldsymbol{\mu}_{t+1}, \Sigma_{t+1}) = \arg\min_{\boldsymbol{\mu},\Sigma} D_{KL}(\mathcal{N}(\boldsymbol{\mu},\Sigma), \mathcal{N}(\boldsymbol{\mu}_t, \Sigma_t))$$

$$+ \frac{1}{2\gamma}\ell^2(\boldsymbol{\mu}; (\mathbf{x}_t, y_t)) + \frac{1}{2\gamma}\mathbf{x}_t^T \Sigma_t \mathbf{x}_t$$

where $\ell^2(\boldsymbol{\mu}; (\mathbf{x}_t, y_t)) = (\max\{0, 1 - y_t(\boldsymbol{\mu} \cdot \mathbf{x}_t)\})^2$ and $\gamma$ is a regularization parameter. The optimization has a closed-form solution similar with CW of (3), but different updating coefficients:

$$\alpha_t = \ell(\boldsymbol{\mu}_t; (\mathbf{x}_t, y_t))\beta_t, \beta_t = \frac{1}{\mathbf{x}_t^T \Sigma_t \mathbf{x}_t + \gamma}$$

## 3. Soft Confidence-Weighted Learning

In this section we present a new online learning method that aims to address the limitation of the CW and AROW learning. Following the same problem settings of the Confidence-Weighted learning, we assume the weight vector $\mathbf{w}$ follows the Gaussian distribution with the mean vector $\boldsymbol{\mu}$ and the covariance matrix $\Sigma$. Notice that the probability constraint in the CW learning, i.e., $Pr_{\mathbf{w} \sim \mathcal{N}(\boldsymbol{\mu}, \Sigma)}[y_t(\mathbf{w} \cdot \mathbf{x}_t) \geq 0] \geq \eta$ can be rewritten as

$$y_t(\boldsymbol{\mu} \cdot \mathbf{x}_t) \geq \phi \sqrt{\mathbf{x}_t^\top \Sigma \mathbf{x}_t},$$

where $\phi = \Phi^{-1}(\eta)$. Further, we introduce a loss function as follows:

$$\ell^\phi(\mathcal{N}(\boldsymbol{\mu}, \Sigma); (\mathbf{x}_t, y_t)) = \max\left(0, \phi\sqrt{\mathbf{x}_t^\top \Sigma \mathbf{x}_t} - y_t \boldsymbol{\mu} \cdot \mathbf{x}_t\right)$$

It is easy to verify that satisfying the probability constraint (i.e., $y_t(\boldsymbol{\mu} \cdot \mathbf{x}_t) \geq \phi\sqrt{\mathbf{x}_t^\top \Sigma \mathbf{x}_t}$ for any $\phi > 0$) is equivalent to satisfying $\ell^\phi(\mathcal{N}(\boldsymbol{\mu}, \Sigma); (\mathbf{x}_t, y_t)) = 0$. Therefore, the optimization problem of the original CW can be re-written as follows

$$(\boldsymbol{\mu}_{t+1}, \Sigma_{t+1}) = \arg\min_{\boldsymbol{\mu}, \Sigma} D_{KL}(\mathcal{N}(\boldsymbol{\mu}, \Sigma) \| \mathcal{N}(\boldsymbol{\mu}_t, \Sigma_t))$$
$$s.t.\ \ell^\phi(\mathcal{N}(\boldsymbol{\mu}, \Sigma); (\mathbf{x}_t, y_t)) = 0,\ \phi > 0$$

The original CW learning method employs a very aggressive updating strategy by changing the distribution as much as necessary to satisfy the constraint imposed by the current example. Although it results in the rapid learning effect, it could force to wrongly change the parameters of the distribution dramatically when handling a mislabeled instance. Such undesirable property makes the original CW algorithm performs poorly in many real-world applications with relatively large noise.

To overcome the above limitation of the CW learning problem, we propose a Soft Confidence-Weighted (SCW) learning method, which aims to soften the aggressiveness of the CW updating strategy. The idea of the SCW learning is inspired by the variants of PA algorithms (PA-I and PA-II) and the adaptive margin. In particular, we formulate the optimization of SCW for learning the soft-margin classifiers as follows:

$$(\boldsymbol{\mu}_{t+1}, \Sigma_{t+1}) = \arg\min_{\boldsymbol{\mu}, \Sigma} D_{KL}(\mathcal{N}(\boldsymbol{\mu}, \Sigma) \| \mathcal{N}(\boldsymbol{\mu}_t, \Sigma_t))$$
$$+ C\ell^\phi(\mathcal{N}(\boldsymbol{\mu}, \Sigma); (\mathbf{x}_t, y_t)) \quad (4)$$

where $C$ is a parameter to tradeoff the passiveness and aggressiveness. We denoted the above formulation of the Soft Confidence-Weighted algorithm, as "SCW-I" for short. Similar to the variant of PA, we can also modify the above formulation by employing a squared penalty, leading to the second formulation of SCW learning (denoted as "SCW-II" for short):

$$(\boldsymbol{\mu}_{t+1}, \Sigma_{t+1}) = \arg\min_{\boldsymbol{\mu}, \Sigma} D_{KL}(\mathcal{N}(\boldsymbol{\mu}, \Sigma) \| \mathcal{N}(\boldsymbol{\mu}_t, \Sigma_t))$$
$$+ C\ell^\phi(\mathcal{N}(\boldsymbol{\mu}, \Sigma); (\mathbf{x}_t, y_t))^2 \quad (5)$$

For the optimization of SCW-I, the following proposition gives the closed-form solution.

**Proposition 1.** *The closed-form solution of the optimization problem (4) is expressed as follows:*

$$\boldsymbol{\mu}_{t+1} = \boldsymbol{\mu}_t + \alpha_t y_t \Sigma_t \mathbf{x}_t, \Sigma_{t+1} = \Sigma_t - \beta_t \Sigma_t \mathbf{x}_t^T \mathbf{x}_t \Sigma_t$$

*where the updating coefficients are as follows:*

$$\alpha_t = \min\{C, \max\{0, \frac{1}{v_t \zeta}(-m_t \psi + \sqrt{m_t^2 \frac{\phi^4}{4} + v_t \phi^2 \zeta})\}\}$$

$$\beta_t = \frac{\alpha_t \phi}{\sqrt{u_t} + v_t \alpha_t \phi}$$

where $u_t = \frac{1}{4}(-\alpha_t v_t \phi + \sqrt{\alpha_t^2 v_t^2 \phi^2 + 4v_t})^2$, $v_t = \mathbf{x}_t^T \Sigma_t \mathbf{x}_t$, $m_t = y_t(\boldsymbol{\mu}_t \cdot \mathbf{x}_t)$, $\phi = \Phi^{-1}(\eta)$, $\psi = 1 + \frac{\phi^2}{2}$ and $\zeta = 1 + \phi^2$.

Similarly, the following proposition gives the closed-form solution to the optimization of SCW-II.

**Proposition 2.** *The closed-form solution of the optimization problem (5) is:*

$$\boldsymbol{\mu}_{t+1} = \boldsymbol{\mu}_t + \alpha_t y_t \Sigma_t \mathbf{x}_t, \Sigma_{t+1} = \Sigma_t - \beta_t \Sigma_t \mathbf{x}_t^T \mathbf{x}_t \Sigma_t$$

*The updating coefficients are as follows:*

$$\alpha_t = \max\{0, \frac{-(2m_t n_t + \phi^2 m_t v_t) + \gamma_t}{2(n_t^2 + n_t v_t \phi^2)}\}$$

$$\beta_t = \frac{\alpha_t \phi}{\sqrt{u_t} + v_t \alpha_t \phi}$$

where $\gamma_t = \phi\sqrt{\phi^2 m_t^2 v_t^2 + 4n_t v_t(n_t + v_t \phi^2)}$, and $n_t = v_t + \frac{1}{2C}$.

The detailed proofs of Proposition 1 and 2 can be found in Appendix section. Finally, Algorithm 1 summarizes the proposed SCW-I and SCW-II algorithms.

## 4. Analysis and Discussions

We first give an overview about the comparison of the proposed SCW methods with respect to several existing first-order and second-order online learning algorithms, followed by the discussions on the nonlinear extension and the bound analysis.

**Algorithm 1** SCW learning algorithms (**SCW**)
  **INPUT:** parameters $C > 0$, $\eta > 0$.
  **INITIALIZATION:** $\boldsymbol{\mu}_0 = (0, \ldots, 0)^\top$, $\Sigma_0 = I$.
  **for** $t = 1, \ldots, T$ **do**
    Receive an example $\mathbf{x}_t \in \mathbb{R}^d$;
    Make prediction: $\hat{y}_t = sgn(\boldsymbol{\mu}_{t-1} \cdot \mathbf{x}_t)$;
    Receive true label $y_t$;
    suffer loss $\ell^\phi\bigl(\mathcal{N}(\boldsymbol{\mu}_{t-1}, \Sigma_{t-1}); (\mathbf{x}_t, y_t)\bigr)$;
    **if** $\ell^\phi\bigl(\mathcal{N}(\boldsymbol{\mu}_{t-1}, \Sigma_{t-1}); (\mathbf{x}_t, y_t)\bigr) > 0$ **then**
      $\boldsymbol{\mu}_{t+1} = \boldsymbol{\mu}_t + \alpha_t y_t \Sigma_t \mathbf{x}_t$, $\Sigma_{t+1} = \Sigma_t - \beta_t \Sigma_t \mathbf{x}_t^T \mathbf{x}_t \Sigma_t$
      where $\alpha_t$ and $\beta_t$ are computed by either Proposition 1 (SCW-I) or Proposition 2 (SCW-II);
    **end if**
  **end for**

### 4.1. Comparison with the existing methods

Following the study of AROW, we qualitatively examine the properties of different algorithms in Table 1. Unlike the previous second-order algorithms, the proposed SCW algorithm enjoys all the four salient properties. In particular, SCW improves over the original CW algorithm by adding the capability to handle the non-separable cases, and improves over AROW by adding the adaptive margin property. To the best of our knowledge, SCW is the first second-order online learning that holds all the four properties.

Table 1. Property comparison of online algorithms.

| Algorithm | Large Margin | Confidence | Non-Separable | Adaptive Margin |
|---|---|---|---|---|
| PA | Yes | No | Yes | No |
| SOP | No | Yes | Yes | No |
| IELLIP | No | Yes | Yes | No |
| CW | Yes | Yes | No | Yes |
| AROW | Yes | Yes | Yes | No |
| NHERD | Yes | Yes | Yes | No |
| NAROW | Yes | Yes | Yes | No |
| SCW | Yes | Yes | Yes | Yes |

### 4.2. Extension to Nonlinear Cases

Similar to other linear online learning methods, the proposed SCW learning can be extended to nonlinear cases. The following lemma shows the possibility of extending the proposed SCW algorithms to nonlinear cases using kernel tricks.

**Lemma 1.** *(Representer Theorem)*
*The mean $\boldsymbol{\mu}_i$ and covariance $\Sigma_i$ parameters computed by the soft confidence weight algorithm can be written as linear combinations of the input vectors with coefficients that depend only on inner products of input vectors, i.e.,*

$$\Sigma_i = \sum_{p,q=1}^{i-1} \pi_{p,q}^{(i)} \mathbf{x}_p \mathbf{x}_q^T + aI, \quad \boldsymbol{\mu}_i = \sum_p^{i-1} \nu_p^{(i)} \mathbf{x}_p$$

*where* $\nu_i^{(i)} = 1$ *and* $\nu_p^{(i+1)} = \nu_p^{(i)} + \alpha_i y_i \sum_q^{i-1} \pi_{p,q}^{(i)} \mathbf{x}_q^T \mathbf{x}_i$ *for* $p < i$, *and* $\pi_{p,q}^{(i+1)} = -\beta_i \sum_{r,s} \pi_{p,r}^{(i)} \pi_{s,q}^{(i)} \mathbf{x}_r^T \mathbf{x}_s + \pi_{p,q}^{(i)}$, $\pi_{p,i}^{(i)} = \pi_{i,p}^{(i)} = -\beta_i \sum_{p,r}^{i-1} \pi_{p,r}^{(i)}(\mathbf{x}_r^T \mathbf{x}_i)$, $\pi_{i,i}^{(i+1)} = -\beta_i$.

The above lemma can be proved by induction similar to the proof in (Crammer et al., 2008).

### 4.3. Analysis of the Loss Bound

Our analysis begins with the definition of confidence loss, which is used in (Crammer et al., 2008). The loss is a function of the margin $m_i$ normalized by $\sqrt{v}$, i.e., $\tilde{m}_i = \frac{m_i}{\sqrt{v_i}}$. We modified the confidence loss in (Crammer et al., 2008) as an upper-bounded loss by:

$$\ell_{\phi_i}(\tilde{m}_i) = \begin{cases} 0 & \tilde{m}_i \geq \phi \\ \min\{f_\phi(\tilde{m}_i), \frac{C^2(1+\phi^2)v_i}{\phi^2}\} & \tilde{m}_i < \phi \end{cases}$$

where $f_\phi(\tilde{m}) = \frac{(-\tilde{m}\psi + \sqrt{\tilde{m}^2 \frac{\phi^4}{4} + \phi^2 \zeta})^2}{\phi^2 \zeta}$. It is easy to see that the loss $\ell_\phi(\tilde{m})$ holds the properties of Lemma 5 in (Crammer et al., 2008) for SCW-I.

We have the following loss bound.

**Theorem 1.** *Let $(x_1, y_1)...(x_n, y_n)$ be an input sequence for SCW-I. Assume there exist $\boldsymbol{\mu}^*$ and $\Sigma^*$ such that for all $i$ for which the algorithm made an update($\alpha_i > 0$),*

$$\boldsymbol{\mu}^{*T} x_i y_i \geq \boldsymbol{\mu}_{i+1}^T x_i y_i, \quad x_i^T \Sigma^* x_i \leq x_i^T \Sigma_{i+1} x_i$$

*Then the following bound holds:*

$$\sum_i \ell_{\phi_i}(\tilde{m}_i) \leq \sum_i (\alpha_i)^2 v_i$$
$$\leq \frac{(1+\phi^2)}{\phi^2}(-\log \det \Sigma^* + \mathrm{Tr}(\Sigma^*) + \boldsymbol{\mu}^{*T} \Sigma_{n+1}^{-1} \boldsymbol{\mu}^* - d)$$

The above theorem can be proved by applying Lemma 7 and property 6 in Lemma 5 in (Crammer et al., 2008). If we let $\ell_{\phi_i}(\tilde{m}_i)$ upper bound the $0-1$ loss by choosing an appropriate $C$, then our mistake number is also bounded by

$$\frac{(1+\phi^2)}{\phi^2}(-\log \det \Sigma^* + \mathrm{Tr}(\Sigma^*) + \boldsymbol{\mu}^{*T} \Sigma_{n+1}^{-1} \boldsymbol{\mu}^* - d).$$

## 5. Empirical Evaluation

### 5.1. Datasets and compared algorithms

We adopt a variety of datasets from different domains:

- synthetic data: we generated this data set by the method described in (Crammer et al., 2008), which is used to examine the effectiveness of second-order algorithms. Following (Crammer et al., 2009b), we also generated another version with 0.1 noise to examine the robustness of second-order algorithms.
- Digital recognition: we use two benchmarks: "USPS" [1] and "MNIST" [2]. For binary classification, we choose "1" vs "all" for "USPS", and "1" vs "2" for "MNIST".
- Face data: we use the MIT-CBCL face imags [3].
- Machine Learning datasets: we randomly choose several public machine learning datasets from [4].

Table 2 shows the statistics of the list of datasets used.

Table 2. List of datasets used in the experiments.

| dataset | # training examples | # features |
| --- | --- | --- |
| splice | 1000 | 60 |
| svmguide3 | 1243 | 21 |
| Synthetic data | 5000 | 20 |
| MITface | 6977 | 361 |
| usps1vsall | 7291 | 256 |
| mushrooms | 8124 | 112 |
| mnist1vs2 | 14867 | 784 |
| w7a | 24692 | 300 |
| codrna | 59535 | 8 |
| ijcnn1 | 141691 | 22 |
| covtype | 581012 | 54 |

We compare our methods with various online learning algorithms, including Perceptron (Rosenblatt, 1958), PA (Crammer et al., 2006), ROMMA (Li & Long, 1999) and its aggressive version agg-ROMMA, Second-Order Perceptron (Cesa-Bianchi et al., 2005), Confidence Weighted Learning (Crammer et al., 2008), Improved Ellipsoid Method for Online Learning(IELLIP) (Yang et al., 2009), AROW (Crammer et al., 2009b), Normal HERD (NHERD) (Crammer & D.Lee, 2010) and NAROW (Orabona & Crammer, 2010). Following the similar parameter setting methods in (Dredze et al., 2008) and (Crammer et al., 2009b). The parameter $r$ in AROW, paramter $b$ in NAROW and parameters $C$ in PA-I, PA-II, NHERD, SCW-I and SCW-II are all determined by cross validation to select the best one from $\{2^{-4}, 2^{-3}, ..., 2^3, 2^4\}$, the parameter $\eta$ in CW, SCW-I and SCW-II are determined by cross validation to select the best one from

[1] http://www-i6.informatik.rwth-aachen.de/~keysers/usps.html
[2] http://yann.lecun.com/exdb/mnist/
[3] http://cbcl.mit.edu/software-datasets/FaceData2.html
[4] http://www.csie.ntu.edu.tw/~cjlin/libsvmtools/datasets/

$\{0.5, 0.55, ..., 0.9, 0.95\}$, the parameter $b$ in IELLIP is determined by cross validation to select the best one from $\{0.1, 0.2, ..., 0.9\}$. After the best parameters are determined, all the experiments were conducted over 20 random permutations for each dataset. All the results were reported by averaging over these 20 runs. We evaluate the performance by three metrics: (i) online cumulative mistake rate, (ii) number of updates (which would be closely related to the potential number of support vectors in kernel extension), and (iii) running time cost.

### 5.2. Experimental Results

Table 3 summarizes the results of our empirical evaluation, where we only show margin-based second-order learning algorithms due to space limitation. For a more complete comparison, please refer to our supplemental material. The **bold** elements indicate the best performance with paired t-test at 95% significance level. We can draw several observations as follows.

First of all, by examining the overall mistakes, we found that second-order algorithms usually outperforms first-order algorithms, and margin based algorithms usually outperforms non-margin based methods. This shows the efficacy of "Large Margin" and "Confidence" properties for learning better classifiers.

Second, by examining the original CW algorithm, we found that, it significantly outperforms the first-order algorithms (e.g. Perceptron, ROMMA, and PA algorithms) on the synthetic data without noise, but fails to outperform the first-order algorithms on some real-world datasets that often have noisy data. This empirical result verifies the importance of "Handling Non-separable" property in producing robust classifiers when dealing with noisy data.

Further, we found that AROW significantly outperforms CW in many real-world datasets (except minist). However, AROW usually produces considerably more updates and spends more running time than CW. This verifies that the importance of "adaptive margin" property of both CW and SCW to reduce the number of updates as well as the running time.

Moreover, among all the compared algorithms, SCW often achieves the best or close to the best performance in terms of accuracy, number of updates, and running time cost. Finally, Figure 1 shows the online results of 13 algorithms with respect to varied numbers of samples in online learning process. The results again validate the advantages of SCW in both efficacy and efficiency among all the state-of-the-art algorithms.

Table 3. Evaluation of cumulative performance of the proposed SCW and other state-of-the-art algorithms.

| Algorithm | svmguide3 | | | codrna | | |
|---|---|---|---|---|---|---|
| | #Mistakes | #Updates | Time(s) | #Mistakes | #Updates | Time(s) |
| CW | 0.294 ± 0.011 | 702.6 ±13.5 | 0.038 ±0.001 | 0.157 ± 0.040 | 9278.9 ±62.8 | 1.083 ±0.018 |
| NHERD | 0.224 ± 0.012 | 1170.3 ±21.6 | 0.052 ±0.004 | 0.089 ± 0.032 | 32232.9 ±8679.4 | 3.086 ±1.180 |
| AROW | 0.218 ± 0.005 | 1174.7 ±15.7 | 0.045 ±0.001 | 0.066 ± 0.000 | 26055.1 ±328.0 | 2.041 ±0.218 |
| NAROW | 0.308 ± 0.096 | 1229.7 ±8.1 | 0.051 ±0.000 | 0.182 ± 0.054 | 54557.1 ±4935.7 | 6.979 ±1.428 |
| SCW-I | **0.209 ± 0.007** | **540.4 ±13.2** | **0.033 ±0.001** | **0.065 ± 0.000** | 7328.8 ±326.1 | **0.952 ±0.021** |
| SCW-II | **0.213 ± 0.008** | 954.2 ±50.9 | 0.044 ±0.001 | 0.066 ± 0.000 | 12070.3 ±438.0 | 1.183 ±0.047 |

| Algorithm | splice | | | usps "1" vs "all" | | |
|---|---|---|---|---|---|---|
| | #Mistakes | #Updates | Time(s) | #Mistakes | #Updates | Time(s) |
| CW | 0.271 ± 0.009 | 555.4 ±9.6 | 0.066 ±0.001 | 0.013 ± 0.001 | 493.7 ±21.4 | 2.723 ±0.100 |
| NHERD | 0.245 ± 0.010 | 805.6 ±22.3 | 0.092 ±0.015 | 0.014 ± 0.001 | 2421.5 ±225.4 | 11.555 ±1.003 |
| AROW | 0.241 ± 0.006 | 741.5 ±24.9 | 0.077 ±0.002 | 0.012 ± 0.001 | 1449.0 ±132.9 | 7.004 ±0.564 |
| NAROW | 0.269 ± 0.015 | 717.9 ±35.2 | 0.085 ±0.017 | 0.018 ± 0.003 | 2153.8 ±251.0 | 10.293 ±1.184 |
| SCW-I | **0.229 ± 0.006** | 541.1 ±8.9 | 0.065 ±0.002 | 0.012 ± 0.001 | 385.5 ±9.7 | **2.221 ±0.049** |
| SCW-II | 0.240 ± 0.010 | **479.80 ±12.4** | **0.050 ±0.001** | **0.011 ± 0.001** | **385.0 ±10.1** | 2.221 ±0.054 |

| Algorithm | ijcnn1 | | | w7a | | |
|---|---|---|---|---|---|---|
| | #Mistakes | #Updates | Time(s) | #Mistakes | #Updates | Time(s) |
| CW | 0.093 ± 0.001 | 30678.0 ±146.9 | 4.844 ±0.142 | 0.104 ± 0.000 | **2432.2 ±48.9** | 17.166 ±0.398 |
| NHERD | 0.084 ± 0.001 | 85104.4 ±4283.4 | 25.014 ±3.075 | 0.101 ± 0.001 | 12348.4 ±378.5 | 79.163 ±2.388 |
| AROW | 0.081 ± 0.000 | 73082.1 ±1272.3 | 16.959 ±1.199 | 0.099 ± 0.001 | 10233.0 ±246.5 | 65.264 ±1.502 |
| NAROW | 0.099 ± 0.020 | 105937.2 ±8231.1 | 39.643 ±6.898 | 0.108 ± 0.001 | 23666.6 ±179.0 | 150.379 ±1.174 |
| SCW-I | **0.058 ± 0.002** | 10561.5 ±704.5 | **2.450 ±0.073** | **0.097 ± 0.000** | 4118.6 ±23.7 | **14.852 ±0.190** |
| SCW-II | 0.072 ± 0.003 | 21792.1 ±3840.6 | 3.823 ±0.568 | 0.099 ± 0.001 | 5634.8 ±78.2 | 24.557 ±0.493 |

| Algorithm | mnist "1" vs "2" | | | MITface | | |
|---|---|---|---|---|---|---|
| | #Mistakes | #Updates | Time(s) | #Mistakes | #Updates | Time(s) |
| CW | 0.012 ± 0.000 | 856.8 ±26.7 | 67.803 ±1.643 | 0.028 ± 0.001 | 835.8 ±19.4 | 8.944 ±0.207 |
| NHERD | 0.108 ± 0.010 | 5258.7 ±415.0 | 335.360 ±24.026 | 0.025 ± 0.001 | 3316.7 ±201.5 | 33.258 ±1.998 |
| AROW | 0.036 ± 0.001 | 4519.0 ±241.0 | 288.433 ±14.286 | 0.027 ± 0.001 | 1884.8 ±134.4 | 19.037 ±1.346 |
| NAROW | 0.038 ± 0.002 | 5819.1 ±356.2 | 372.040 ±21.925 | 0.031 ± 0.002 | 2389.8 ±215.4 | 24.185 ±2.158 |
| SCW-I | **0.011 ± 0.001** | 868.3 ±22.8 | 68.507 ±1.399 | 0.025 ± 0.001 | **756.1 ±14.6** | **8.131 ±0.175** |
| SCW-II | **0.011 ± 0.001** | **742.9 ±34.0** | **60.901 ±2.081** | **0.024 ± 0.001** | 774.0 ±20.2 | 8.320 ±0.207 |

| Algorithm | mushrooms | | | covtype | | |
|---|---|---|---|---|---|---|
| | #Mistakes | #Updates | Time(s) | #Mistakes | #Updates | Time(s) |
| CW | **0.002 ± 0.000** | 315.8 ±18.0 | 0.289 ±0.005 | 0.405 ± 0.001 | 389870.8 ±2278.9 | 879.788 ±9.014 |
| NHERD | **0.002 ± 0.001** | 3724.4 ±448.3 | 1.245 ±0.125 | 0.259 ± 0.002 | 521225.6 ±12104.9 | 1166.939 ±34.727 |
| AROW | **0.002 ± 0.000** | 1815.0 ±185.5 | 0.662 ±0.056 | 0.243 ± 0.000 | 531187.0 ±455.4 | 1193.212 ±4.858 |
| NAROW | **0.002 ± 0.000** | 3340.8 ±386.0 | 1.138 ±0.109 | 0.367 ± 0.009 | 546704.0 ±8814.8 | 1269.775 ±45.753 |
| SCW-I | **0.002 ± 0.000** | 327.6 ±21.1 | 0.288 ±0.006 | **0.233 ± 0.000** | **238415.6 ±1917.1** | **264.846 ±2.501** |
| SCW-II | **0.002 ± 0.000** | **152.7 ±5.7** | **0.241 ±0.010** | 0.239 ± 0.000 | 451193.3 ±3783.0 | 881.011 ±9.477 |

| Algorithm | synthetic data | | | synthetic data with 0.1 noise | | |
|---|---|---|---|---|---|---|
| | #Mistakes | #Updates | Time(s) | #Mistakes | #Updates | Time(s) |
| CW | **0.017 ± 0.001** | **262.9 ±6.7** | **0.071 ±0.000** | 0.293 ± 0.005 | 2811.1 ±33.9 | 0.129 ±0.007 |
| NHERD | 0.120 ± 0.016 | 3202.9 ±292.6 | 0.134 ±0.006 | 0.208 ± 0.017 | 4114.6 ±150.7 | 0.156 ±0.003 |
| AROW | 0.026 ± 0.003 | 2128.0 ±167.3 | 0.084 ±0.004 | **0.133 ± 0.003** | 4122.8 ±60.7 | 0.136 ±0.001 |
| NAROW | 0.101 ± 0.019 | 3302.0 ±315.0 | 0.125 ±0.009 | 0.236 ± 0.024 | 4242.8 ±171.8 | 0.154 ±0.007 |
| SCW-I | 0.018 ± 0.001 | 317.9 ±8.6 | **0.071 ±0.000** | 0.135 ± 0.002 | **1373.1 ±38.9** | **0.094 ±0.006** |
| SCW-II | 0.020 ± 0.001 | 326.0 ±6.6 | **0.071 ±0.000** | 0.145 ± 0.005 | 2138.9 ±211.7 | 0.111 ±0.008 |

## 6. Conclusion

This paper proposed the Soft Confidence-Weighted (SCW) learning, a new second-order online learning method with state-of-the-art empirical performance. Unlike the existing second-order algorithms, SCW enjoys all the four properties: (i) large margin training, (ii) confidence weighting, (iii) adaptive margin, and (iv) capability of handling non-separable data. Empirically, we found the proposed SCW algorithms perform significantly better than the original CW algorithm, and outperform the state-of-the-art AROW algorithm for most cases in terms of both accuracy and efficiency. Future work will conduct more in-depth analysis of the mistake bounds and its multi-class extension (Crammer et al., 2009a).

## Appendix: Proof of Proposition 1 and 2

*Proof.* First, when $\ell^\phi(\mathcal{N}(\boldsymbol{\mu}_t, \Sigma_t); (\mathbf{x}_t, y_t)) = 0$, it is easy to see the solution is valid. When $\ell^\phi(\mathcal{N}(\boldsymbol{\mu}_t, \Sigma_t); (\mathbf{x}_t, y_t)) > 0$, it is easy to see the optimization problem is equivalent to

$$D_{KL}(\mathcal{N}(\boldsymbol{\mu}, \Sigma) \| \mathcal{N}(\boldsymbol{\mu}_t, \Sigma_t)) + C\xi,$$
$$s.t. \quad \ell^\phi(\mathcal{N}(\boldsymbol{\mu}, \Sigma); (\mathbf{x}_t, y_t)) \leq \xi \quad and \quad \xi \geq 0$$

Since $\Sigma$ is positive semi-definite (PSD), it can be written as $\Sigma = \Upsilon^2$ to make the optimization with a convex constraint in $\boldsymbol{\mu}$ and $\Upsilon$ simultaneously. But for convenient, we will still use $\Sigma$ instead of $\Upsilon^2$ in the following analysis. The Lagrangian of the above optimization is

$$\begin{aligned}
&\mathcal{L}(\boldsymbol{\mu}, \Sigma, \xi, \tau, \lambda) \\
&= D_{KL}(\mathcal{N}(\boldsymbol{\mu}, \Sigma) \| \mathcal{N}(\boldsymbol{\mu}_t, \Sigma_t)) + C\xi \\
&+ \tau(\phi\sqrt{\mathbf{x}_t^\top \Sigma \mathbf{x}_t} - y_t \boldsymbol{\mu} \cdot \mathbf{x}_t - \xi) - \lambda\xi \\
&= D_{KL}(\mathcal{N}(\boldsymbol{\mu}, \Sigma) \| \mathcal{N}(\boldsymbol{\mu}_t, \Sigma_t)) \\
&+ \xi(C - \tau - \lambda) + \tau(\phi\sqrt{\mathbf{x}_t^\top \Sigma \mathbf{x}_t} - y_t \boldsymbol{\mu} \cdot \mathbf{x}_t) \\
&= \frac{1}{2}\log(\frac{\det \Sigma_t}{\det \Sigma}) + \frac{1}{2}Tr(\Sigma_t^{-1}\Sigma) + \frac{1}{2}(\boldsymbol{\mu}_t - \boldsymbol{\mu})^\top \Sigma_t^{-1}(\boldsymbol{\mu}_t - \boldsymbol{\mu}) \\
&- \frac{d}{2} + \xi(C - \tau - \lambda) + \tau(\phi\sqrt{\mathbf{x}_t^\top \Sigma \mathbf{x}_t} - y_t \boldsymbol{\mu} \cdot \mathbf{x}_t)
\end{aligned}$$

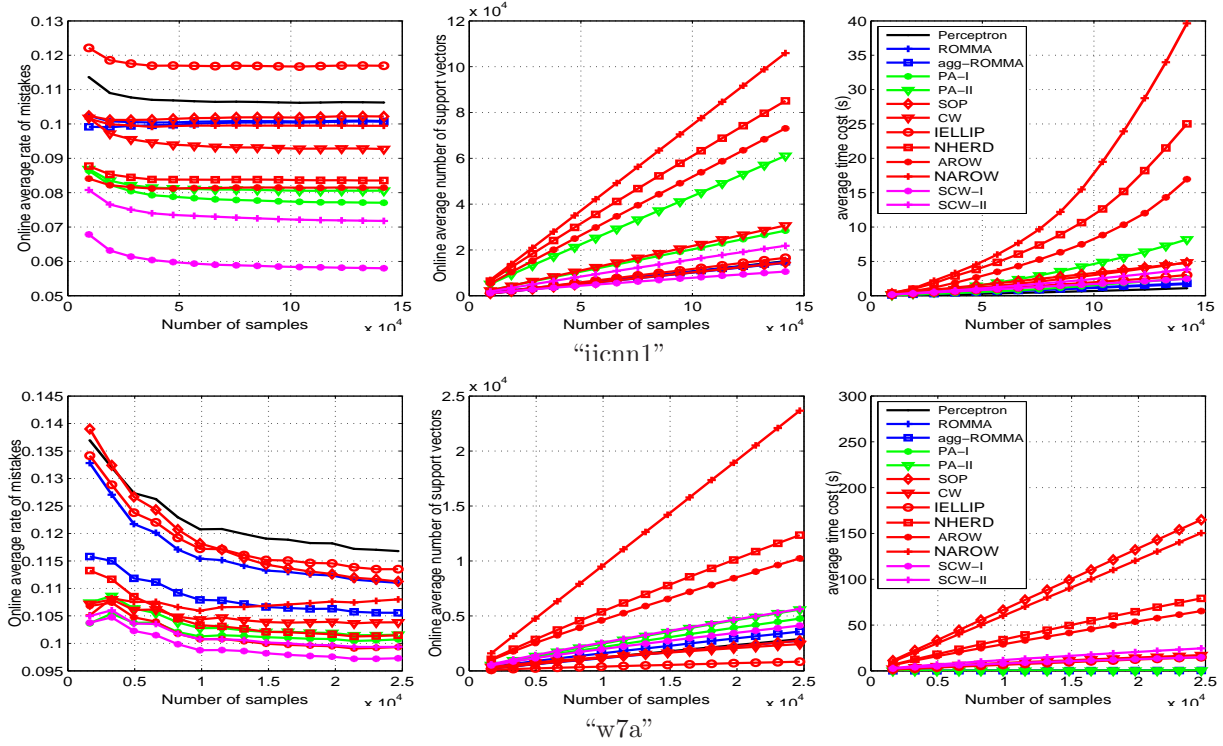

Figure 1. Evaluation of online performance of the proposed SCW and other state-of-the-art algorithms.

where $\tau \geq 0$ and $\lambda \geq 0$ are Lagrange multipliers. We now find the minimum of the Lagrangian with respect to the primal variables $\boldsymbol{\mu}$, $\Sigma$ and $\xi$.

$$\frac{\partial \mathcal{L}}{\partial \boldsymbol{\mu}} = \Sigma_t^{-1}(\boldsymbol{\mu} - \boldsymbol{\mu}_t) + \tau(-y_t \mathbf{x}_t) = 0 \Rightarrow \boldsymbol{\mu} = \boldsymbol{\mu}_t + \tau y_t \Sigma_t \mathbf{x}_t$$

$$\frac{\partial \mathcal{L}}{\partial \Sigma} = 0 \Rightarrow \Sigma_{t+1}^{-1} = \Sigma_t^{-1} + \tau \phi \frac{\mathbf{x}_t \mathbf{x}_t^\top}{\sqrt{\mathbf{x}_t^\top \Sigma_{t+1} \mathbf{x}_t}}$$

and $C - \tau - \lambda = 0$, so $\tau = C - \lambda \leq C$, thus $\tau \in [0, C]$
The KKT conditions for the optimization are:

$$\phi \sqrt{\mathbf{x}_t^\top \Sigma \mathbf{x}_t} - y_t \boldsymbol{\mu} \cdot \mathbf{x}_t - \xi \leq 0, -\xi \leq 0, \tau, \lambda \geq 0$$

$$\tau(\phi \sqrt{\mathbf{x}_t^\top \Sigma \mathbf{x}_t} - y_t \boldsymbol{\mu} \cdot \mathbf{x}_t - \xi) = 0, \lambda \xi = 0$$

**Case 1.** $\tau \neq 0$
As $\tau(\phi \sqrt{\mathbf{x}_t^\top \Sigma \mathbf{x}_t} - y_t \boldsymbol{\mu} \cdot \mathbf{x}_t - \xi) = 0$ implies $(\phi \sqrt{\mathbf{x}_t^\top \Sigma \mathbf{x}_t} - y_t \boldsymbol{\mu} \cdot \mathbf{x}_t - \xi) = 0$, the KKT conditions are simplified:

$$-\xi \leq 0, \tau > 0, \lambda \geq 0$$

$$\phi \sqrt{\mathbf{x}_t^\top \Sigma \mathbf{x}_t} - y_t \boldsymbol{\mu} \cdot \mathbf{x}_t - \xi = 0, \lambda \xi = 0$$

**Sub-case 1.1.** $\lambda \neq 0$
When $\lambda \neq 0$, $\lambda \xi = 0$, implies $\xi = 0$. The KKT conditions are simplified as

$$\tau > 0, \lambda > 0, \xi = 0, \phi \sqrt{\mathbf{x}_t^\top \Sigma \mathbf{x}_t} - y_t \boldsymbol{\mu} \cdot \mathbf{x}_t = 0$$

Finally, we have the following:

$$\Sigma_{t+1} = \left(\Sigma_t^{-1} + \tau \phi \frac{\mathbf{x}_t \mathbf{x}_t^\top}{\sqrt{\mathbf{x}_t^\top \Sigma_{t+1} \mathbf{x}_t}}\right)^{-1}$$

$$= \Sigma_t - \Sigma_t \mathbf{x}_t \left(\frac{\tau \phi}{\sqrt{\mathbf{x}_t^\top \Sigma_{t+1} \mathbf{x}_t} + \tau \phi \mathbf{x}_t^\top \Sigma_t \mathbf{x}_t}\right) \mathbf{x}_t^\top \Sigma_t$$

Let $u_t = \mathbf{x}_t^\top \Sigma_{t+1} \mathbf{x}_t$, $v_t = \mathbf{x}_t^\top \Sigma_t \mathbf{x}_t$, $m_t = y_t(\boldsymbol{\mu}_t \cdot \mathbf{x}_t)$, multiplying by $\mathbf{x}_t^\top$ (left) and $\mathbf{x}_t$ (right), we get $u_t = v_t - v_t(\frac{\tau \phi}{\sqrt{u_t} + \tau \phi v_t}) v_t$, which can be used to solve $u_t$:

$$\sqrt{u_t} = \frac{-\tau \phi v_t + \sqrt{\tau^2 \phi^2 v_t^2 + 4 v_t}}{2}.$$

And $\phi \sqrt{\mathbf{x}_t^\top \Sigma \mathbf{x}_t} - y_t \boldsymbol{\mu} \cdot \mathbf{x}_t = 0$ implies $\phi \sqrt{u_t} - m_t - \tau v_t = 0$. Thus, $\phi \frac{-\tau \phi v_t + \sqrt{\tau^2 \phi^2 v_t^2 + 4 v_t}}{2} - m_t - \tau v_t = 0$, which can be rearranged as: $v_t^2(1 + \phi^2)\tau^2 + 2 m_t v_t(1 + \frac{\phi^2}{2})\tau + (m_t^2 - \phi^2 v_t)$. The larger root is then

$$\tau = \frac{-m_t v_t(1 + \frac{\phi^2}{2}) + \sqrt{\Delta}}{v_t^2(1 + \phi^2)},$$

where $\Delta = m_t^2 v_t^2(1 + \frac{\phi^2}{2})^2 - v_t^2(1 + \phi^2)(m_t^2 - \phi^2 v_t)$.
If $\tau \in (0, C)$, then $\lambda = C - \tau \in (0, C)$.
**Sub-case 1.2.** $\lambda = 0$
$C - \tau - \lambda = 0$ implies $\tau = C$. The KKT conditions can be simplified as:

$$-\xi \leq 0, \tau = C, \lambda = 0, \phi \sqrt{\mathbf{x}_t^\top \Sigma \mathbf{x}_t} - y_t \boldsymbol{\mu} \cdot \mathbf{x}_t - \xi = 0$$

We thus have:

$$\phi\sqrt{\mathbf{x}_t^\top \Sigma \mathbf{x}_t} - y_t \boldsymbol{\mu} \cdot \mathbf{x}_t$$
$$= [\phi\frac{-\tau\phi v_t + \sqrt{\tau^2\phi^2 v_t^2 + 4v_t}}{2} - m_t - \tau v_t]|_{\tau=C} = \xi \geq 0$$

It is easy to verify that

$$f'(\tau) = \frac{-\phi^2 v_t}{2} + \frac{\phi^3 v_t^2 \tau}{2\sqrt{\tau^2\phi^2 v_t^2 + 4v_t}} - v_t = 0$$

has no solution on $[0, +\infty)$ and $f'(0) = \frac{-\phi^2 v_t}{2} - v_t < 0$. As a result, $f'(\tau) < 0$, $\tau \in [0, +\infty)$, which implies $f(\tau)$ is decreasing on $[0, +\infty)$.

$$f(C) \geq 0 = f(\theta)$$

where $\theta = \frac{-m_t v_t(1+\frac{\phi^2}{2}) + \sqrt{m_t^2 v_t^2(1+\frac{\phi^2}{2})^2 - v_t^2(1+\phi^2)(m_t^2 - \phi^2 v_t)}}{v_t^2(1+\phi^2)}$, which thus implies $C \leq \theta$.

**Case 2.** $\tau = 0$

When $\tau = 0$, since $C - \tau - \lambda = 0$, $\lambda = C$, the KKT condtions are simplified as

$$\phi\sqrt{\mathbf{x}_t^\top \Sigma \mathbf{x}_t} - y_t\boldsymbol{\mu} \cdot \mathbf{x}_t \leq 0, \tau = 0, \lambda = C, \xi = 0$$

Thus, $\boldsymbol{\mu}_{t+1} = \boldsymbol{\mu}_t$ and $\Sigma_{t+1} = \Sigma_t$; as a result, $\phi\sqrt{\mathbf{x}_t^\top \Sigma_t \mathbf{x}_t} - y_t\boldsymbol{\mu}_t \cdot \mathbf{x}_t \leq 0$, which contradicts with $\ell^\phi\big(\mathcal{N}(\boldsymbol{\mu}, \Sigma); (\mathbf{x}_t, y_t)\big) > 0$.

For SCW-II, the Lagrangian of the optimization is

$$\mathcal{L}(\boldsymbol{\mu}, \Sigma, \xi, \tau, \lambda) = D_{KL}\big(\mathcal{N}(\boldsymbol{\mu}, \Sigma) \| \mathcal{N}(\boldsymbol{\mu}_t, \Sigma_t)\big) + C\xi^2$$
$$+ \tau(\phi\sqrt{\mathbf{x}_t^\top \Sigma \mathbf{x}_t} - y_t\boldsymbol{\mu} \cdot \mathbf{x}_t - \xi) - \lambda\xi$$
$$= D_{KL}\big(\mathcal{N}(\boldsymbol{\mu}, \Sigma) \| \mathcal{N}(\boldsymbol{\mu}_t, \Sigma_t)\big)$$
$$+ \xi(C\xi - \tau - \lambda) + \tau(\phi\sqrt{\mathbf{x}_t^\top \Sigma \mathbf{x}_t} - y_t\boldsymbol{\mu} \cdot \mathbf{x}_t)$$
$$= \frac{1}{2}\log(\frac{\det \Sigma_t}{\det \Sigma}) + \frac{1}{2}Tr(\Sigma_t^{-1}\Sigma) + \frac{1}{2}(\boldsymbol{\mu}_t - \boldsymbol{\mu})^\top \Sigma_t^{-1}(\boldsymbol{\mu}_t - \boldsymbol{\mu})$$
$$- \frac{d}{2} + \xi(C\xi - \tau - \lambda) + \tau(\phi\sqrt{\mathbf{x}_t^\top \Sigma \mathbf{x}_t} - y_t\boldsymbol{\mu} \cdot \mathbf{x}_t)$$

where $\tau \geq 0$ and $\lambda \geq 0$ are Lagrange multipliers. We now find the minimum of the Lagrangian with respect to the primal variables $\boldsymbol{\mu}$, $\Sigma$ and $\xi$.

$$\frac{\partial \mathcal{L}}{\partial \boldsymbol{\mu}} = \Sigma_t^{-1}(\boldsymbol{\mu} - \boldsymbol{\mu}_t) + \tau(-y_t\mathbf{x}_t) = 0 \Rightarrow \boldsymbol{\mu} = \boldsymbol{\mu}_t + \tau y_t \Sigma_t \mathbf{x}_t$$

$$\frac{\partial \mathcal{L}}{\partial \Sigma} = 0 \Rightarrow \Sigma_{t+1}^{-1} = \Sigma_t^{-1} + \tau\phi\frac{\mathbf{x}_t \mathbf{x}_t^\top}{\sqrt{\mathbf{x}_t^\top \Sigma_{t+1} \mathbf{x}_t}}$$

and $2C\xi - \tau - \lambda = 0$, so $\xi = \frac{\tau + \lambda}{2C}$. The KKT conditions for the optimization are:

$$\phi\sqrt{\mathbf{x}_t^\top \Sigma \mathbf{x}_t} - y_t\boldsymbol{\mu} \cdot \mathbf{x}_t - \xi \leq 0$$
$$\xi \geq 0, \tau \geq 0, \lambda \geq 0$$
$$\tau(\phi\sqrt{\mathbf{x}_t^\top \Sigma \mathbf{x}_t} - y_t\boldsymbol{\mu} \cdot \mathbf{x}_t - \xi) = 0$$
$$\lambda\xi = 0$$

The rest proof is similar to that of SCW-I. $\square$

## Acknowledgments
This work was in part supported by Singapore MOE tier 1 project (RG33/11) and Microsoft Research project (M4060936).

## References

Cesa-Bianchi, Nicolò, Conconi, Alex, and Gentile, Claudio. A second-order perceptron algorithm. *SIAM J. Comput.*, 34(3):640–668, 2005.

Crammer, Koby and D.Lee, Daniel. Learning via gaussian herding. In *NIPS*, pp. 345–352, 2010.

Crammer, Koby, Dekel, Ofer, Keshet, Joseph, Shalev-Shwartz, Shai, and Singer, Yoram. Online passive-aggressive algorithms. *Journal of Machine Learning Research*, 7:551–585, 2006.

Crammer, Koby, Dredze, Mark, and Pereira, Fernando. Exact convex confidence-weighted learning. In *NIPS*, pp. 345–352, 2008.

Crammer, Koby, Dredze, Mark, and Kulesza, Alex. Multi-class confidence weighted algorithms. In *EMNLP*, pp. 496–504, 2009a.

Crammer, Koby, Kulesza, Alex, and Dredze, Mark. Adaptive regularization of weight vectors. In *NIPS*, pp. 345–352, 2009b.

Dredze, Mark, Crammer, Koby, and Pereira, Fernando. Confidence-weighted linear classification. In *ICML*, pp. 264–271, 2008.

Duchi, John C., Hazan, Elad, and Singer, Yoram. Adaptive subgradient methods for online learning and stochastic optimization. *Journal of Machine Learning Research*, 12:2121–2159, 2011.

Jin, Rong, Hoi, Steven C. H., and Yang, Tianbao. Online multiple kernel learning: Algorithms and mistake bounds. In *ALT*, pp. 390–404, 2010.

Li, Bin, Zhao, Peilin, Hoi, Steven C. H., and Gopalkrishnan, Vivekanand. Pamr: Passive aggressive mean reversion strategy for portfolio selection. *Machine Learning*, 87(2):221–258, 2012.

Li, Yi and Long, Philip M. The relaxed online maximum margin algorithm. In *NIPS*, pp. 498–504, 1999.

Orabona, Francesco and Crammer, Koby. New adaptive algorithms for online classification. In *NIPS*, pp. 1840–1848, 2010.

Rosenblatt, Frank. The perceptron: A probabilistic model for information storage and organization in the brain. *Psych. Rev.*, 7:551–585, 1958.

Yang, Liu, Jin, Rong, and Ye, Jieping. Online learning by ellipsoid method. In *ICML*, pp. 145, 2009.

Zhao, Peilin, Hoi, Steven C. H., and Jin, Rong. Double updating online learning. *Journal of Machine Learning Research*, 12:1587–1615, 2011a.

Zhao, Peilin, Hoi, Steven C. H., Jin, Rong, and Yang, Tianbao. Online auc maximization. In *ICML*, pp. 233–240, 2011b.